%% file: main.tex
\documentclass[10pt,twocolumn,letterpaper]{article}

\usepackage{iccv}              


\definecolor{iccvblue}{rgb}{0.21,0.49,0.74}
\usepackage[pagebackref,breaklinks,colorlinks,allcolors=iccvblue]{hyperref}

\usepackage{colortbl}

\def\paperID{00018} 
\def\confName{ICCV}
\def\confYear{2025}

\title{Hybrid Generative Fusion for Efficient and Privacy-Preserving  \\ Face Recognition Dataset Generation}

\author{
	Feiran Li\textsuperscript{1,2}\hspace{1em} Qianqian Xu\textsuperscript{3,}\thanks{Corresponding authors}\hspace{1em} Shilong Bao\textsuperscript{4}\hspace{1em} Boyu Han\textsuperscript{3,4} \hspace{1em} Zhiyong Yang\textsuperscript{4} \hspace{1em} Qingming Huang\textsuperscript{4,3,5,*} \\
	{\textsuperscript{1}Institute of Information Engineering, Chinese Academy of Sciences} \\
	{\textsuperscript{2}School of Cyber Security, University of Chinese Academy of Sciences} \\
    {\textsuperscript{3}Institute of Computing Technology, Chinese Academy of Sciences} \\
    {\textsuperscript{4}School of Computer Science and Technology, University of Chinese Academy of Sciences} \\
    {\textsuperscript{5}BDKM, University of Chinese Academy of Sciences} \\
	{\tt\small lifeiran@iie.ac.cn \hspace{2em} \{xuqianqian, hanboyu23z\}@ict.ac.cn} \\ 
    {\tt\small \{baoshilong, yangzhiyong21, qmhuang\}@ucas.ac.cn }
}

\begin{document}
\maketitle
\input{sec/0_abstract}    
\input{sec/1_intro}
\input{sec/2_method}
\input{sec/3_exp}

\input{sec/4_con}
{
    \small
    \bibliographystyle{ieeenat_fullname}
    \bibliography{main}
}

\end{document}

%% file: sec/0_abstract.tex
\begin{abstract}
In this paper, we present our approach to the DataCV ICCV Challenge, which centers on building a high-quality face dataset to train a face recognition model. The constructed dataset must not contain identities overlapping with any existing public face datasets. 
To handle this challenge, we begin with a thorough cleaning of the baseline HSFace dataset, identifying and removing mislabeled or inconsistent identities through a Mixture-of-Experts (MoE) strategy combining face embedding clustering and GPT-4o-assisted verification. We retain the largest consistent identity cluster and apply data augmentation up to a fixed number of images per identity. 
To further diversify the dataset, we generate synthetic identities using Stable Diffusion with prompt engineering. As diffusion models are computationally intensive, we generate only one reference image per identity and efficiently expand it using Vec2Face, which rapidly produces 49 identity-consistent variants. This hybrid approach fuses GAN-based and diffusion-based samples, enabling efficient construction of a diverse and high-quality dataset.
To address the high visual similarity among synthetic identities, we adopt a curriculum learning strategy by placing them early in the training schedule, allowing the model to progress from easier to harder samples. Our final dataset contains 50 images per identity, and all newly generated identities are checked with mainstream face datasets to ensure no identity leakage. 
Our method achieves \textbf{1st place} in the competition, and experimental results show that our dataset improves model performance across 10K, 20K, and 100K identity scales. Code is available at~\url{https://github.com/Ferry-Li/datacv_fr}.

\end{abstract}

%% file: sec/1_intro.tex
\section{Introduction}
\label{sec:intro}
Recent advances in face recognition (FR) have been largely driven by deep learning models~\cite{shao2023facial,hua2024reconboost} trained on large-scale, labeled datasets. However, due to growing privacy, ethical, and legal concerns, the use of real facial identities ~\cite{shao2023identity,shao2025facial,qinmixbridge,wu2024image,wu2023consistency} is increasingly restricted or infeasible. The \textit{DataCV ICCV Face Recognition Dataset Construction Challenge} poses this issue by introducing the task of creating high-quality training datasets~\cite{wu2024vec2face,kim2023dcface,boutros2023idiff,deandres2024frcsyn,shahreza2024sdfr,wu2025generative,wu2023logical,wu2023should} that do not include any real-world identities. The objective is to train a fixed face recognition model with only the synthetic identities, aiming to achieve performance on test sets that is comparable to or better than models trained on real identities.

While synthetic data offers a promising alternative, it introduces several notable challenges:

\begin{enumerate}
    \item \textbf{Identity Consistency:} Generative models, particularly diffusion-based approaches, are designed to maximize diversity and realism. However, this makes it difficult to generate multiple distinct images that consistently represent the same synthetic identity~\cite{papantoniou2024arc2face,zhang2023adding}. Such inconsistency can confuse the training process and reduce recognition accuracy.

    \item \textbf{Intra-Class Variation:} For robust generalization, FR models must handle variations~\cite{qiu2021synface,boutros2022sface,shao2024joint,shao2025mol,han2024aucseg,liu2024not,wei2025decoupling} in pose, lighting, expression, and background. Synthetic datasets often lack this natural diversity, especially when samples are generated from a single reference image or through narrowly constrained pipelines.

    \item \textbf{Identity Leakage:} The dataset must not contain any identities overlapping with public real-world datasets. This is a strict requirement of the competition and is challenging to enforce when using pretrained generative models or external data sources without explicit identity controls.

    \item \textbf{Computational Cost:} High-quality image generation using diffusion models is computationally expensive, especially when scaling to thousands of identities with dozens of images each. This limits the feasibility of large-scale synthesis without significant computing resources.
\end{enumerate}

\begin{figure*}[t]
    \centering
    \includegraphics[width=\linewidth]{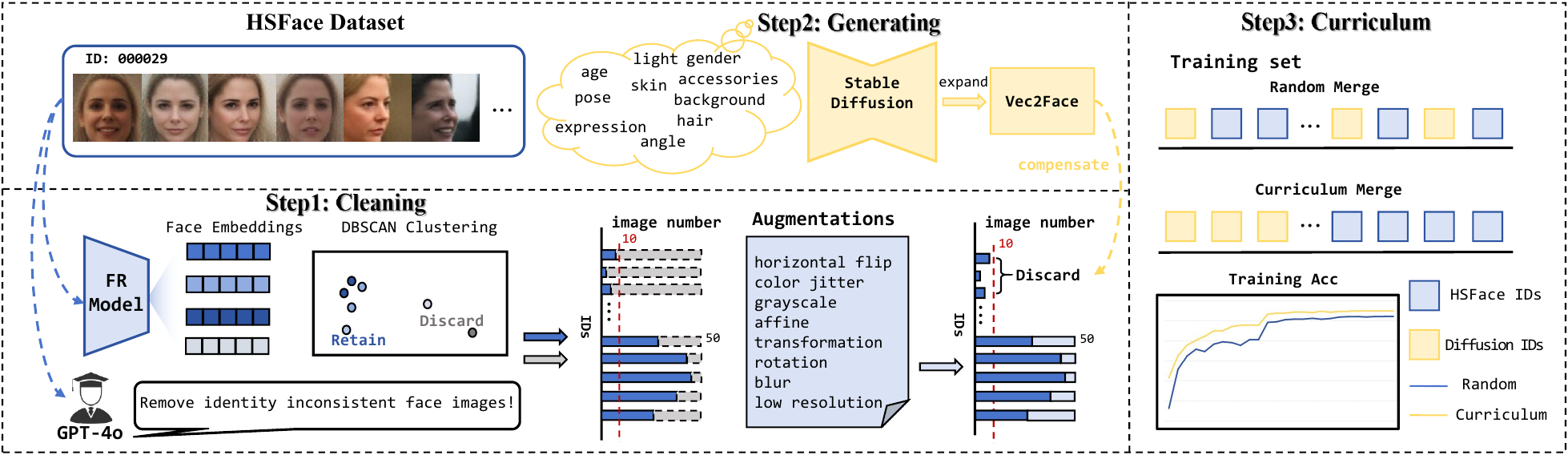}
    \caption{An overview of our method. We first perform data cleaning (Step 1) using a face recognition model and GPT-4o validation, followed by a DBSCAN clustering to retain only consistent face images. Data augmentation is applied to restore the number of images per identity to 50. We then generate additional identities (Step 2) using Stable Diffusion and Vec2Face, introducing controlled variations in attributes such as pose, expression, and lighting. Finally, in Step 3, we construct the training set using a curriculum learning strategy that merges diffusion-generated and HSFace identities in a staged manner, improving convergence and generalization compared to random mixing.}
    \label{fig:pipeline}
\end{figure*}

To address these challenges, we propose a hybrid approach that combines \textbf{dataset cleaning}, \textbf{identity synthesis}, and \textbf{curriculum-based training structuring}. Our method begins by refining the baseline \textit{HSFace} dataset~\cite{wu2024vec2face}. While HSFace provides a strong starting point, we observe that some identities exhibit label noise or intra-identity inconsistency. To improve the dataset quality, we adopt a Mixture-of-Experts (MoE)~\cite{lepikhingshard,huaopenworldauc} strategy that integrates automated face embedding clustering and GPT-4o-assisted~\cite{achiam2023gpt} validation. For each identity, we retain the largest consistent image cluster and apply data augmentation to ensure 50 samples per identity. Identities with excessive inconsistency (i.e., more than 80\% of images removed) are discarded.

To make up for the vacancy after the cleaning and ensure diversity, we generate new identities using Stable Diffusion~\cite{rombach2022high} guided by prompt engineering~\cite{lione2025,huaopenworldauc,wu2025prompt}. For each generated identity, a single reference image is created, and 49 additional samples are synthesized using Vec2Face~\cite{wu2024vec2face}, resulting in visually consistent image sets. However, these generated samples tend to be overly similar. To mitigate the risk of overfitting, we apply a curriculum learning~\cite{bengio2009curriculum,wu2024top} strategy, positioning these easy (low intra-class variance) identities at the beginning of the training dataset, followed by harder (high intra-class variance) identities~\cite{bao2024improved,bao2025aucpro,wu2024image,wu2025image} from the cleaned HSFace set. This gradual exposure improves the model’s ability to generalize.

Importantly, we verify that all synthetic identities do not overlap with any known real individuals, fully satisfying the competition’s identity non-leakage requirement. Our final dataset is constructed at three scales: 10K, 20K, and 100K identities, each with 50 images, resulting in a total of 500K, 1M, and 5M synthetic face images, respectively. Evaluated by the organizers, our submission achieves \textbf{1st place} on the competition leaderboard across all three scales, demonstrating the effectiveness of our synthetic dataset construction strategy in enhancing face recognition performance under fixed training conditions.

%% file: sec/2_method.tex
\section{Method}
Our approach is designed to construct a high-quality synthetic training dataset that enables a fixed face recognition model to achieve strong performance without using any real identities. The pipeline consists of three main components: dataset cleaning, identity generation, and curriculum-based data structuring.

\subsection{Overview}
Figure~\ref{fig:pipeline} illustrates the overall pipeline of our method. We begin by refining the provided \textit{HSFace} dataset~\cite{wu2024vec2face} to remove noisy or inconsistent identities using a Mixture-of-Experts (MoE) strategy. Simultaneously, we generate new synthetic identities using Stable Diffusion and Vec2Face to increase diversity and scale. The resulting identities are structured using a curriculum learning strategy to improve training efficiency and model generalization. Finally, the cleaned and generated identities are combined into datasets of 10K, 20K, and 100K identities, each with 50 images per identity.

\subsection{Dataset Cleaning}

\begin{figure}
    \centering
    \includegraphics[width=\linewidth]{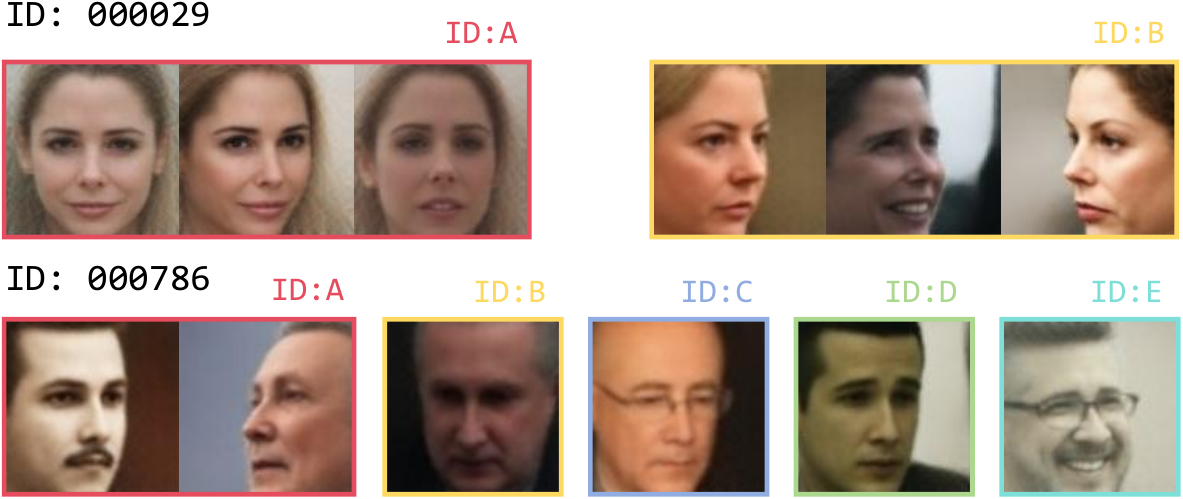}
    \caption{Images with inconsistent identities exist in HSFace.}
    \label{fig:incon_ID}
\end{figure}

We start with the \textit{HSFace} dataset provided by the organizers. While it provides a strong baseline, we observe that some identities exhibit label noise and intra-identity inconsistency, which can negatively affect training, as shown in Figure~\ref{fig:incon_ID} where \texttt{ID:000029} contains images subject to two identities, and \texttt{ID:000786} contains images subject to more than four identites. To improve dataset quality, we apply a \textbf{Mixture-of-Experts (MoE)} cleaning strategy that includes:
\begin{itemize}
    \item \textbf{Face embedding clustering} using a pretrained FR model to identify inconsistent images within identities.
    \item \textbf{GPT-4o-assisted validation} to resolve ambiguous identity clusters using both semantic and visual cues.
\end{itemize}

Specifically, we extract face embeddings for each identity using the baseline-provided model \texttt{hsface300K.pth}. These embeddings are clustered using the DBSCAN~\cite{ester1996density} algorithm with cosine similarity as the distance metric. For each identity, we retain only the largest cluster, which we assume to be the most consistent subset. Identities with less than 20\% of their original samples remaining after clustering are discarded to ensure label purity.

For ambiguous cases, where image quality is low or visual consistency is difficult to assess, we employ GPT-4o for multimodal identity validation. We generate a composite image containing all face samples of a given identity, each annotated with a unique index. This image is submitted to GPT-4o along with a carefully designed prompt asking the model to identify which samples do not belong to the same identity as the majority. A prompt template is listed as follows:
\begin{quote}
``You will receive an image consisting of several face photos arranged in a grid, where each face has a numeric ID label below it. Please identify all the face images that do not belong to the same person as the majority. Your answer should only include the numeric IDs of the outlier faces, separated by commas (e.g., 001,005,012). Do not include any additional text, punctuation, or whitespace. Do not return an empty response."
\end{quote}
To ensure response validity, we apply a regular expression filter to verify that GPT-4o's output matches the expected format. This multimodal validation mechanism is particularly effective in resolving borderline cases where visual similarity alone is insufficient.

By combining these experts, we are able to construct a high-quality cleaned subset of HSFace, which serves as one of the key components in our final training dataset.
To maintain a consistent number of 50 images per identity, we apply data augmentation to the remaining valid samples after cleaning. Specific details of the augmentation pipeline are provided in Section~\ref{sec:details}.

\begin{figure}[t]
    \centering
    \includegraphics[width=\linewidth]{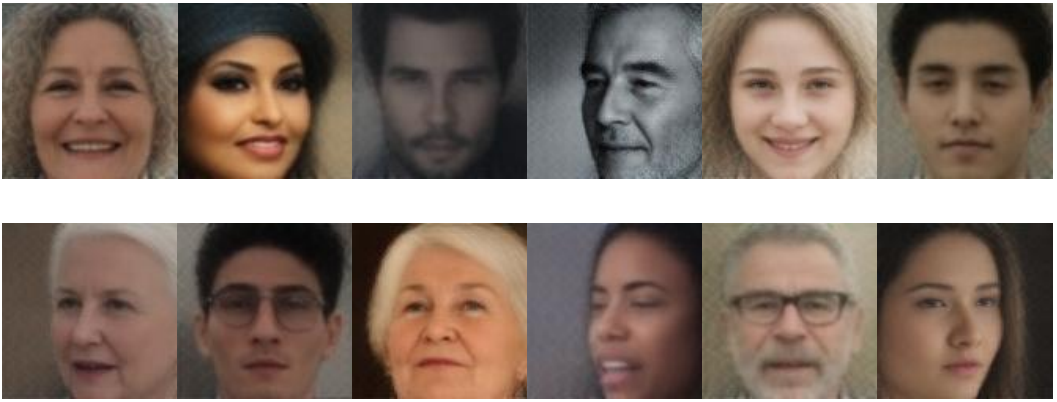}
    \caption{Representative samples generated by the prompts with Stable Diffusion.}
    \label{fig:sd_example}
\end{figure}

\subsection{Identity Generation}
To expand the identity space and compensate for identities removed during the dataset cleaning stage, we construct a large number of synthetic identities using a two-stage pipeline. This process combines Stable Diffusion for identity creation with Vec2Face for identity-consistent sample expansion.

We begin by sampling a prompt from a curated list of hundreds of base identity descriptions. These prompts are carefully designed to specify characteristics such as age, gender, ethnicity, facial structure, hairstyle, and accessories, while ensuring that the generated subject's face is clearly visible and unobstructed. This prompt is passed to a Stable Diffusion XL~\footnote{The model is provided in \url{https://huggingface.co/stabilityai/stable-diffusion-xl-base-1.0/tree/main}.} text-to-image model, which synthesizes one high-resolution portrait image per novel synthetic identity.
To ensure diversity in pose, illumination, and expression across the full identity set, we introduce variation at the prompt level. Each identity prompt is augmented with a secondary description specifying randomly sampled attributes from predefined lists, including head pose (e.g., “facing left”, “tilted right”), expression (e.g., “slight smile”, “serious look”), lighting condition (e.g., “under natural daylight”, “lit by a soft lamp”), camera angle, background, and optional accessories. This prompt augmentation helps maximize the inter-identity diversity of the generated reference images. Representative samples produced through this prompting scheme are shown in Figure~\ref{fig:sd_example}.

Once a reference image is generated by Stable Diffusion, we apply a face detector using the \texttt{insightface.app.FaceAnalysis} toolkit~\cite{guo2020towards} to ensure that the image contains a clearly visible and valid face. If no face is detected, the sample is discarded. Otherwise, the largest detected face region is cropped and used as the canonical identity image.
To efficiently expand each identity into a diverse sample set, we feed the cropped reference image to the Vec2Face model~\cite{wu2024vec2face}, a lightweight decoder that rapidly generates 49 identity-consistent variants. These variations include controlled changes in pose, illumination, and facial expression, while preserving the core identity features. Compared to diffusion-based methods, Vec2Face offers a significant reduction in computational cost, making it well-suited for large-scale dataset generation.

This two-stage pipeline, leveraging Stable Diffusion for high-quality identity initialization and Vec2Face for fast, consistent sample generation, produces a visually coherent identity class of 50 images. It ensures high inter-identity diversity and strong intra-identity consistency, both of which are critical for efficient and effective face recognition model training.
\begin{figure}
    \centering
    \includegraphics[width=\linewidth]{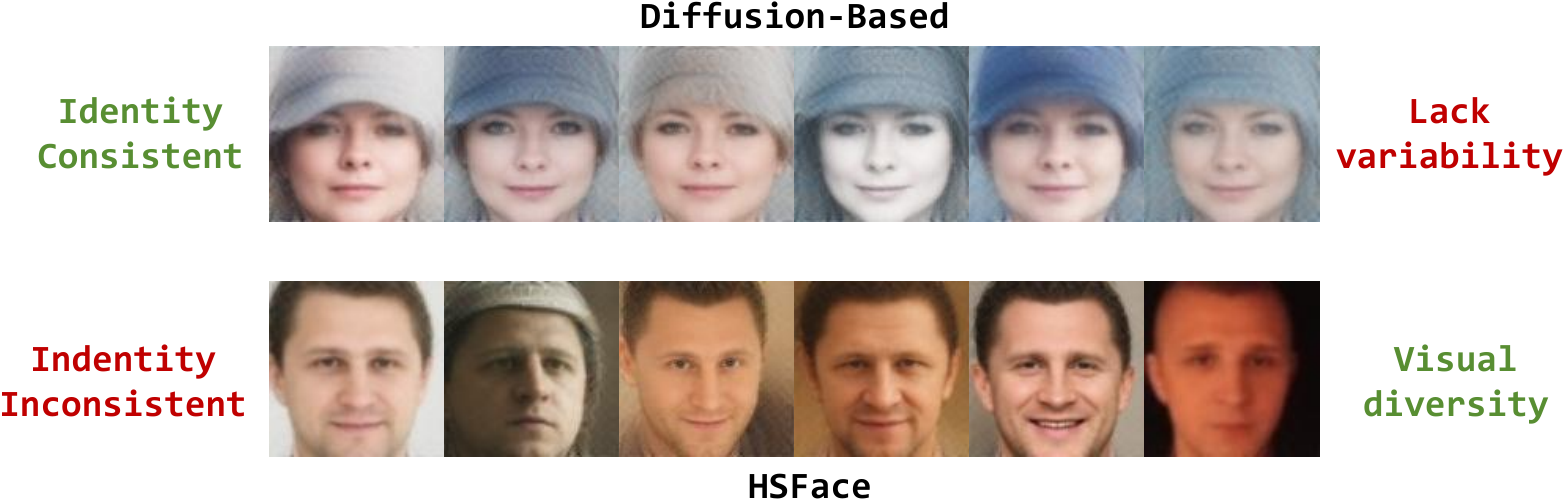}
    \caption{Diffusion-generated images tend to be visually similar and maintain identity consistency, while HSFace samples exhibit greater visual diversity but may include identity inconsistencies.}
    \label{fig:easy_vs_hard}
\end{figure}
\subsection{Curriculum-Based Data Structuring}

While the generated identities exhibit strong intra-class consistency, they tend to lack the variability, such as natural changes in expression, pose, and lighting. In contrast, the cleaned subset of the HSFace dataset contains more visually diverse and realistic samples, but also introduces greater intra-class variation and residual noise, making it more challenging for the model to learn from. This contrast is illustrated in Figure~\ref{fig:easy_vs_hard}, which presents representative examples from both the generated and the HSFace datasets.

To mitigate this imbalance and improve generalization, we adopt a curriculum learning strategy that organizes the training data in order of increasing difficulty. The model is first exposed to synthetic images generated via Stable Diffusion and Vec2Face, which offer clean and identity-consistent samples. As training progresses, it encounters more diverse and complex examples from the cleaned HSFace dataset. This curriculum enables the model to first establish robust identity representations from homogeneous data before adapting to higher variability. Empirically, we find that this progression leads to faster convergence and better recognition performance, particularly under constrained training budgets.

In practice, we discard identities from the original HSFace subset that have too few valid samples, as identified during the data cleaning process. For each removed identity, a replacement is randomly selected from a pool of synthetic identities, each contributing a complete set of images (e.g., 50 samples). This ensures that the total number of identities remains unchanged while preserving consistent intra-class structure.
The final dataset is constructed by placing all selected synthetic identities at the beginning of the training set, followed by the remaining cleaned HSFace identities. The model is first exposed to high-quality, identity-consistent synthetic samples before encountering the more diverse and challenging real-world data. Identity grouping is preserved throughout, and no shuffling is applied to the dataset construction. This simple yet effective structure facilitates stable convergence and contributes significantly to improved recognition accuracy.

%% file: sec/3_exp.tex
\section{Experiments}

\subsection{Implementation Details}
\label{sec:details}

\paragraph{Hardware and Framework.}
All experiments are conducted on a single NVIDIA RTX 4090 GPU. The implementation is based on PyTorch.

\begin{table*}[t]
  \centering
  \caption{Test results with 10K-scale training set. The best result is shown in \textbf{bold}.}
  \label{tab:10K_exp}
    \begin{tabular}{ccccc}
    \toprule
    \textbf{Participants} & \textbf{Average ACC (\%)} & \textbf{ACC 1 (\%)} & \textbf{ACC 2 (\%)} & \textbf{ACC 3 (\%)} \\
    \midrule
    nicolo.didomenico & 77.26 & 84.84 & 77.67 & 69.26 \\
    mnogolikomus & 82.83 & 88.21 & 92.52 & 67.77 \\
    Lin0  & 83.20 & 88.50 & 92.58 & 68.53 \\
    f10942093 & 83.25 & 88.97 & 93.12 & 67.67 \\
    GZ\_Xu & 83.45 & 88.59 & 92.63 & 69.14 \\
    XLW   & 83.51 & 91.20 & 91.03 & 68.30 \\
    anjith2006 & 83.66 & 91.26 & 89.42 & \textbf{70.30} \\
    Qile\_Xu & 84.31 & 88.69 & \textbf{94.35} & 69.91 \\
    \rowcolor[rgb]{ .89,  .949,  .851} \textbf{Ours} & \textbf{84.37} & \textbf{91.53} & 92.62 & 68.97 \\
    \bottomrule
    \end{tabular}%
    
  \vspace{1.5em}
  
  \caption{Test results with 20K-scale training set. The best result is shown in \textbf{bold}.}
  \label{tab:20K_exp}
    \begin{tabular}{ccccc}
    \toprule
    \textbf{Participants} & \textbf{Average ACC (\%)} & \textbf{ACC 1 (\%)} & \textbf{ACC 2 (\%)} & \textbf{ACC 3 (\%)} \\
    \midrule
    nicolo.didomenico & 76.80 & 81.47 & 78.98 & 69.95 \\
    mnogolikomus & 83.03 & 88.21 & 92.32 & 68.55 \\
    f10942093 & 84.49 & 89.86 & \textbf{93.85} & 69.76 \\
    anjith2006 & 85.13 & 92.23 & 91.20 & \textbf{71.97} \\
    \rowcolor[rgb]{ .89,  .949,  .851} \textbf{Ours} & \textbf{85.43} & \textbf{92.69} & 93.40 & 70.20 \\
    \bottomrule
    \end{tabular}%
    
  \vspace{1.5em}
  
  \caption{Test results with 100K-scale training set. The best result is shown in \textbf{bold}.}
  \label{tab:100K_exp}
    \begin{tabular}{ccccc}
    \toprule
    \textbf{Participants} & \textbf{Average ACC (\%)} & \textbf{ACC 1 (\%)} & \textbf{ACC 2 (\%)} & \textbf{ACC 3 (\%)} \\
    \midrule
    nicolo.didomenico & 77.96 & 83.20  & 80.98 & 69.68 \\
    anjith2006 & 85.74 & 92.19 & 92.88 & 72.14 \\
    \rowcolor[rgb]{ .89,  .949,  .851} \textbf{Ours} & \textbf{86.78} & \textbf{93.96} & \textbf{94.17} & \textbf{72.23} \\
    \bottomrule
    \end{tabular}%
  \label{tab:100k_exp}%
\end{table*}%

\paragraph{Feature Extraction.}
We use the pretrained model \texttt{hsface300.pth}~\footnote{The pretrained model is provided in \url{https://huggingface.co/BooBooWu/Vec2Face/tree/main/fr_weights}.} as our feature extractor during data cleaning and clustering. This model is provided in the Vec2Face baseline. No training is performed, except during the final fixed training required by the competition.

\paragraph{Base Datasets.}
We build our dataset based on three datasets: HSFace-10K, HSFace-20K, and HSFace-100K~\footnote{Base datasets are provided in \url{https://huggingface.co/datasets/BooBooWu/Vec2Face/tree/main/HSFaces}.}, which contain 10K, 20K, and 100K identities, respectively. Each identity contains at most 50 images.

\paragraph{Multimodal Language Model.}
All prompt generation and description filtering are performed using the GPT-4o API, which enables high-quality textual reasoning and attribute manipulation for identity prompts and caption validation.

\paragraph{Data Cleaning Parameters.}
Identities with fewer than 10 remaining images after filtering are discarded. To ensure intra-identity visual consistency, we apply DBSCAN clustering using cosine distance on features. The similarity threshold is dynamically adjusted between 0.3 and 0.9 to ensure that the largest cluster retains between 50\% and 80\% of the total images for each identity.

\paragraph{Data Augmentation.}
After data cleaning, we apply a consistent data augmentation pipeline to all remaining training images. The primary goal of this step is to restore the number of images per identity to a fixed count of 50, compensating for samples removed during filtering and clustering. In addition to replenishing the dataset, the augmentation strategy introduces controlled variations in pose, illumination, and resolution, thereby enhancing model robustness while preserving identity consistency. The full augmentation sequence is as follows:

\begin{itemize}
  \item Random horizontal flip with probability 0.5
  \item Random color jitter (brightness, contrast, saturation, hue) with probability 0.8
  \item Random grayscale conversion with probability 0.2
  \item Random affine transformation (rotation up to 10°, translation up to 5\%, scaling from 0.95 to 1.05, shear up to 5°) with probability 0.5
  \item Random in-plane rotation (±5°) with probability 0.5
  \item Gaussian blur with kernel size 3 and $\sigma \in [0.1, 2.0]$
  \item Downsampling to low resolution by a factor of 0.5 using a custom \texttt{low\_resolution()} function
\end{itemize}

This augmentation pipeline is applied offline after data cleaning and is used to generate new images that are stored in the final training dataset. As a result, \textbf{no modifications} to the training code or runtime transforms are necessary.

\subsection{Results}
Our submission is evaluated by the competition organizers in the face recognition challenge under three different training scales: 10K, 20K, and 100K identities. Each track corresponds to a fixed-size training set with 50 images per identity and is assessed using three official test subsets (ACC 1, ACC 2, ACC 3). The final ranking is determined by the average accuracy across these three subsets.

As shown in Tables 1, 2, and 3, our method consistently achieves the \textbf{1st place} across all scales. We obtain the highest average accuracy at each scale, \textbf{84.37\%} (10K), \textbf{85.43\%} (20K), and \textbf{86.78\%} (100K), and lead on multiple individual test sets. These results demonstrate the effectiveness and scalability of our curriculum-based training pipeline, high-quality synthetic data generation, and robust identity cleaning process.

%% file: sec/4_con.tex
\section{Conclusion}
This paper addresses the challenge of constructing high-quality synthetic datasets for face recognition without relying on any real-world identities. In response to the requirements of the \textit{DataCV ICCV Face Recognition Dataset Construction Challenge}, we propose a unified pipeline that combines dataset cleaning, identity synthesis, and curriculum-based training.

We first enhance the HSFace dataset by removing noisy and inconsistent samples through embedding-based clustering and GPT-4o-assisted validation. This improves identity consistency while preserving intra-class diversity. To supplement the cleaned dataset, we generate new identities using Stable Diffusion and Vec2Face, ensuring visual coherence and variation across attributes such as pose, lighting, and expression. To guide the training process, we introduce a curriculum learning strategy that orders training data from low to high difficulty based on intra-class variation, allowing the model to gradually adapt and generalize better.

All generated identities are verified to be free of any overlap with real-world individuals, fully satisfying the competition’s identity non-leakage constraint. The resulting datasets, constructed at three different scales, achieve first place across all tracks of the challenge. This demonstrates the effectiveness of our synthetic data construction strategy in training competitive face recognition models under constrained settings.